\documentclass[lettersize,journal]{IEEEtran}
\usepackage{amsmath,amsfonts}
\usepackage{algorithmic}
\usepackage{array}
\usepackage[caption=false,font=normalsize,labelfont=sf,textfont=sf]{subfig}
\usepackage{textcomp}
\usepackage{stfloats}
\usepackage{url}
\usepackage{verbatim}
\usepackage{booktabs}
\usepackage{graphicx}
\usepackage{siunitx}
\usepackage{hyperref}
\usepackage{multirow}
\usepackage{comment}
\usepackage{graphicx}
\usepackage{dcolumn}
\usepackage{bm}
\usepackage{ulem}
\usepackage[dvipsnames]{xcolor}

\def\BibTeX{{\rm B\kern-.05em{\sc i\kern-.025em b}\kern-.08em
    T\kern-.1667em\lower.7ex\hbox{E}\kern-.125emX}}
\usepackage{balance}
\begin{document}

\title{Inverse Design of Snap-Actuated Jumping Robots Powered by Mechanics-Aided Machine Learning}



\author{Dezhong Tong, Zhuonan Hao, Mingchao Liu$^*$, Weicheng Huang$^*$
\thanks{Dezhong Tong is with Department of Material Science and Engineering, University of Michigan, Ann Arbor, 2800 Plymouth Rd, Ann Arbor, MI, 48105 (email: dezhong@umich.edu ).}
\thanks{Zhuonan Hao is with Department of Mechanical and Aerospace Engineering, University of California, Los Angeles, 420 Westwood Plaza, Los Angeles, CA 90095 (email: znhao@g.ucla.edu).}
\thanks{Mingchao Liu is with Department of Mechanical Engineering, University of Birmingham, Birmingham, B15 2TT, UK (email: m.liu.2@bham.ac.uk).}
\thanks{Weicheng Huang is with the School of Engineering, Newcastle University, Newcastle upon Tyne, NE1 7RU, UK (email: weicheng.huang@ncl.ac.uk).}
\thanks{$^*$Corresponding author: Mingchao Liu and Weicheng Huang}
}

\maketitle

\begin{abstract}
Exploring the design and control strategies of soft robots through simulation is highly attractive due to its cost-effectiveness. 
Although many existing models (e.g., finite element analysis) are effective for simulating soft robotic dynamics, there remains a need for a general and efficient numerical simulation approach in the soft robotics community.
In this paper, we develop a discrete differential geometry-based numerical framework to achieve the model-based inverse design of a novel snap-actuated jumping robot.
It is found that the dynamic process of a snapping beam can be either symmetric or asymmetric, such that the trajectory of the jumping robot can be tunable (e.g., horizontal or vertical). 
By employing this novel mechanism of the bistable beam as the robotic actuator, we next propose a physics-data hybrid inverse design strategy for the snap-jump robot with a broad spectrum of jumping capabilities.
We first use the physical engine to study the influences of the robot's design parameters on the jumping capabilities, then generate extensive simulation data to formulate a data-driven inverse design solution. 
The inverse design solution can rapidly explore the combination of design parameters for achieving a target jump, which provides valuable guidance for the fabrication and control of the jumping robot.
The proposed methodology paves the way for exploring the design and control insights of soft robots with the help of simulations.

\end{abstract}


\section{Introduction}
\label{intro}

The development of autonomous robots capable of performing dynamic and complex movements has become a focal point in robotics research. 
In recent years, robots with different types of locomotion at different scales have been created, including crawling, rolling, jumping, flying, and swimming~\cite{calisti2017fundamentals,ng2021locomotion,sun2021soft}. 
Among various locomotion strategies, jumping mechanisms have garnered significant attention due to their efficiency and ability to navigate challenging terrains~\cite{chen2021legless,zhang2020biologically}.
They have shown great potential for diverse applications, ranging from search and rescue operations to freight transportation and planetary exploration~\cite{tsukagoshi2005jumping,kolvenbach2019towards,yang2023morphing}.

\begin{figure*}[t]
\centering
\includegraphics[width=\textwidth]{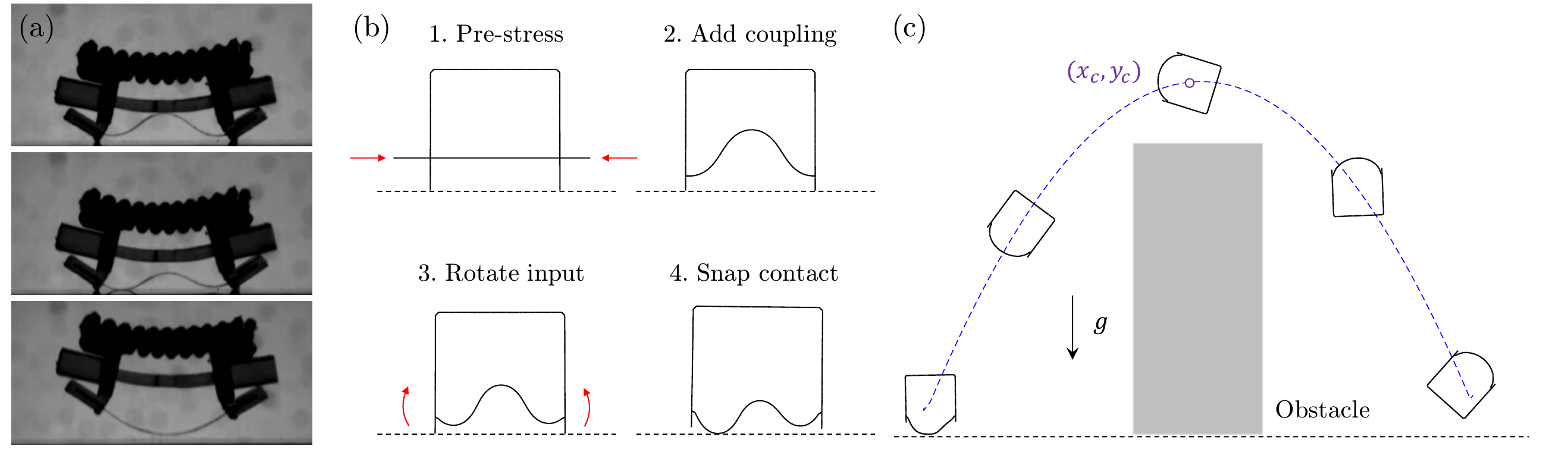}
\caption{Snapshots for the snap-induced jump robot. (a) Snapshots from experiments. (b) Numerical setup. (c) Snapshots from simulation. The experimental figures are from Ref. \cite{wang2023insect}.}
\label{fig:introPlot}
\end{figure*}

Jumping is a common type of locomotion observed in a wide range of biological systems, such as the squat jump in frogs, rabbits, and humans; the spring jump in grasshoppers and fleas; and the snap jump in click beetles and trap-jaw ants~\cite{aura1989biomechanical}. 
These jumping mechanisms provide abundant inspiration for engineers designing bio-mimetic jumping robots \cite{noh2012flea,xu2023design,yun20233}.
Compared to squat jumps and spring jumps, snap-actuated jumping mechanisms offer a more efficient way to store and release energy rapidly by leveraging snap-through instability, enabling animals and insects to achieve high-speed and high-force movements with minimal energy input.
For example, click beetles have demonstrated this efficiency by achieving jump heights over 50 times their body length \cite{bolmin2021nonlinear}.
Mimicking this mechanism, an insect-scale jumping robot has been created \cite{wang2023insect}, capable of powerful jumps (see Fig.~\ref{fig:introPlot}(a)), with several similar examples following \cite{yang2024snapping,guo2024bistable}.

Leveraging the power of snap-through buckling, the jumping robot can usually achieve remarkable locomotion performance, as shown in many prior works~\cite{wang2023insect, guo2024bistable, gorissen2020inflatable}.
However, few studies consider how to tune the buckling to optimize the control and design of such soft robots.
Recent studies suggest that different dynamic transitions can be achieved for a pre-compressed beam by introducing defects at the boundary conditions~\cite{wang2024transient}.
We incorporate this mechanism into the fundamental design principles presented in Ref.~\cite{wang2023insect}.
By tuning the buckling of the pre-compressed beam~\cite{wang2024transient}, which serves as the soft actuator, the robot can jump along a broad spectrum of trajectories.
%

Designing soft robots, particularly those utilizing snap-actuated mechanisms, traditionally relies heavily on iterative testing and empirical adjustments. This approach is both time-consuming and resource-intensive.
Therefore, the use of physics-based models to explore design and control schemes of soft structures in a `sim2real' manner is becoming increasingly prevalent in robotics~\cite{choi2024learning, tong2024sim2real}.
A common approach to modeling soft robots is using Finite Element Analysis (FEA) software (e.g., ANSYS, Abaqus, COMSOL). However, these tools are normally computationally intensive and slow to converge when they involve complex internal and external interactions, making it challenging to optimize design and control schemes directly through simulations~\cite{armanini2023soft}.
To ensure model efficiency, some prior works have developed specific models for soft robots. For example, a simple mathematical model presented in Ref.~\cite{wang2023insect} predicts the jumping performance of an insect-scale snapping robot, showing potential for guiding optimal design.
However, such models are often limited to very specific conditions, like symmetric geometry and actuation, due to their simplicity.
Recently, numerical models based on discrete differential geometry (DDG) have shown great potential for modeling soft structures's behaviors accurately, such as buckling of ribbons, knots and shells~\cite{huang2024exploiting,tong2023snap}, as well as crawling, rolling and swimming robots~\cite{huang2023modeling,huang2020dynamic,huang2022design}. 
Due to the reduced-order expression of soft structures, DDG-based simulations can achieve efficient modeling of soft robots~\cite{qin2024modeling, choi2024dismech} with lower computational costs compared to FEA.
In this work, we adopt the DDG-based framework to study the design and control of a novel snap-actuated jumping robot.

We utilize the DDG-based framework to design an accurate and efficient simulation tool to thoroughly study the influence of design parameters on the proposed robot. 
Additionally, we formulate a novel data-driven inverse design optimization problem to explore the design parameters needed to achieve a target jump. 
This approach provides valuable guidance for the fabrication and control of the jumping robot. 
We demonstrate how to develop design and control strategies for soft robots using physics-based simulations.

The primary contributions of our work are outlined below:
\begin{itemize}
\item We propose a novel design for a snap-actuated jumping robot that leverages the tunable symmetrical/asymmetrical buckling of a pre-compressed beam to achieve various jumps;
\item We develop a robust and accurate physics-based simulation framework to comprehensively analyze the design parameters of the robots;
\item We construct a data-driven model to facilitate the rapid exploration of the required design parameters for the robot to achieve a target jump. This model can later be used in the control of the proposed soft robot.
\end{itemize}

Moreover, we offer simulation videos and release all our code as open-source software.
\footnote{See \href{https://github.com/DezhongT/Jumping\_Robot}{https://github.com/DezhongT/Jumping\_Robot}}

The paper is organized as follows. In Section~\ref{sec:problem}, we describe the design and the relevant parameters of the snap-actuated robot. Section~\ref{sec:numerical} details the physically-based simulation used to study the robot's performance, and Section~\ref{sec:analysis} analyzes the mechanism of the robot's jump, focusing on the snap-through buckling of the pre-compressed beam as the soft actuator. Then, a data-driven approach is depicted in Section~\ref{sec:inverseDesign} for the fast inverse design of the robot, enabling control for a target jump. Section~\ref{sec:conclusion} provides concluding remarks and directions for future work.

\section{Problem formulation}
\label{sec:problem}
Our proposed design utilizes the tunable asymmetric snap-through buckling of a pre-compressed beam as the jumping actuator. Fig.~\ref{fig:mechanicsPlot}(a1) illustrates the schematic of the jumping robot with net mass $m$, which consists of two components: a rigid body frame with horizontal and vertical side lengths $l_1$ and $l_2$, and a deformable pre-compressed beam. The beam, with an original length $L$ and pre-compression (i.e. end-shortening) $\Delta L$, is mounted on the rigid body frame at a height $h$ from the bottom of the robot. The rigid body frame height $l_2$ is fixed while the length is varied as the pre-compression, i.e., $l_{1}= L - \Delta L$. The friction coefficient between the robot and the substrate is denoted as $\mu$.

As depicted in Fig.~\ref{fig:introPlot}(b), rotating the two ends of the pre-compressed beam causes it to rapidly change shape and strike the ground. This snap-through buckling process, triggered by the rotational boundary condition, releases stored energy, generating a substantial interaction force between the robot and the substrate. This force propels the robot into a parabolic jump, as shown in Fig.~\ref{fig:introPlot}(c). For simplicity, we characterize the jump by its highest point of the center of mass $(x_c, y_c)$.

The robot's jumping behavior is influenced by its geometry, material's properties, and the interaction between robot and substrate. In this manuscript, we extensively explore how these parameters affect the snap-through buckling mechanism and the resulting jump with a novel DDG-based physical-based simulation framework. Furthermore, we leverage the simulation data to construct a data-driven approach to efficiently and accurately inverse design the robot, identifying the critical design parameters necessary to achieve the desired jump performance.

\section{Numerical Framework}
\label{sec:numerical}
In this section, we discuss how to incorporate frictional contact with the elastic deformation of solid structures to simulate the proposed snapping-induced jumping robot.
%
First, we employ the DDG-based model to calculate the nonlinear elastic deformations of the soft robot accurately~\cite{huang2020dynamic}.  
Then, the frictional contact between the robot and the rigid substrate is modeled using artificial contact potential~\cite{li2020incremental, tong2023fully}, whose efficiency and physical accuracy have been demonstrated in prior works~\cite{choi2021implicit, tong2023fully}.
Finally, we apply the classical implicit Euler method to formulate the discrete equations of motion \cite{huang2019newmark}.


\subsection{Elastic deformation} 

\begin{figure*}[t]
\centering
\includegraphics[width=1.0\textwidth]{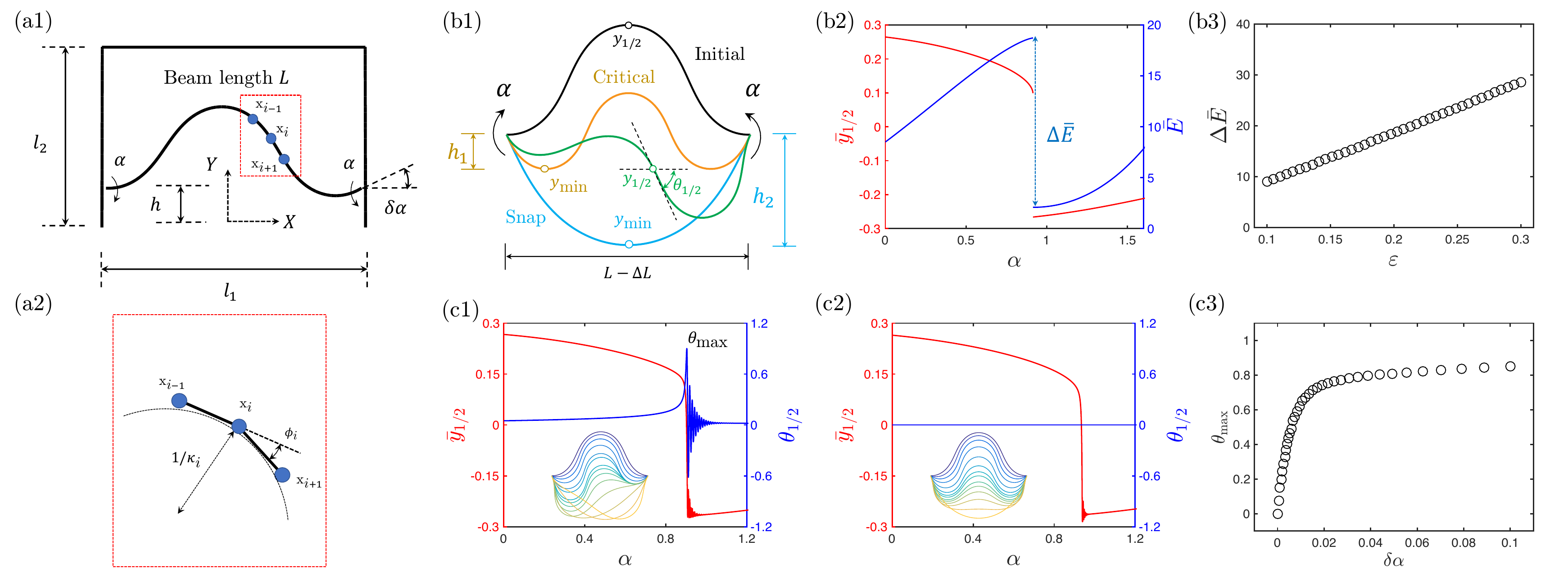}
\caption{Snap bifurcation of a pre-compressed beam under rotational input. (a1) Geometry of the snap-jump robot. (a2) Discrete elements in numerical simulation. (b1) Equilibrium configurations during the snap-through process. (b2) Static bifurcation diagram. (b3) Energy releases as a function of pre-compression ratio $\varepsilon$. (c1) Dynamic snap-through for asymmetric clamped boundary condition. (c2) Dynamic snap-through for perfect symmetric clamped boundary condition. (c3) Maximum midpoint angle during the snap-through as a function of mismatch angle $\delta \alpha$.}
\label{fig:mechanicsPlot}
\end{figure*}

As illustrated in Fig.~\ref{fig:mechanicsPlot}(a1), the geometry of the jumping robot can be discretized into $\mathcal{N}$ nodes, each denoted by $\mathbf x_i \in \mathbb R^{2\times 1}$, for $i \in [0, \mathcal{N})$. Those discrete nodes are connected by a sequence of edges, each denoted by $\mathbf e_i$. 
Therein, a $2\mathcal{N} \times 1$ degrees of freedom (DOF) vector $\mathbf{q} = [\mathbf x_0, \mathbf x_1, ..., \mathbf x_{\mathcal{N}-1}]$ can be constructed to describe the jumping robot's configuration.
With the discretization scheme defined, the formulation of elastic strains and energies can be outlined with the DOF vector $\mathbf q$. The strains of the proposed robot are composed of stretching and bending.

The stretching strain of an edge connecting discrete nodes $\mathbf x_i$ and $\mathbf x_{i+1}$ is given by:
\begin{equation}
\xi_{i} = \frac {|| \mathbf x_{i+1} - \mathbf x_i ||} {|| \hat{\mathbf{e}}_{i} ||} - 1.
\end{equation}
Hereafter, quantities with $\hat{(\cdot)}$ indicates the undeformed status. For example, $||\hat{\mathbf{e}}_{i}||$ is the undeformed length of the edge $\mathbf e_i$. The discrete stretching energy for an edge is:
\begin{equation}
E^s_i = \frac{1}{2} EA \xi_{i}^2 || \hat{\mathbf{e}}_{i} ||,
\end{equation}
where $EA=Ewb$ represents the local stretching stiffness (where $ w$ is the in-plane width and the $b$ is the thickness). The bending strain, which corresponds to the local curvature $\kappa_i$, can be computed using
\begin{equation}
\kappa_i = 2 \frac {\tan(\phi_i/2) }  {\Delta \hat{l}_{i} },
\end{equation}
where $\phi_i$ is the turning angle, as shown in Fig.~\ref{fig:mechanicsPlot} (2a), and $\Delta \hat{l}_{i}$ is the Voronoi length at node $\mathbf x_i$. Then, the formulation of discrete bending energy is
\begin{equation}
E^b_i = \frac{1}{2} EI \left( \kappa_i - \hat{\kappa}_i \right)^{2} \Delta \hat{l}_{i},
\end{equation}
where $EI=Ewb^3/12$ represents the bending stiffness,
and $\hat{\kappa}_i$ represents the natural curvature in the undeformed configuration. 

The total elastic energy of the system is the sum of all discretized elastic energies:
\begin{equation}
E^{\text{ela}} = \sum_{i}^{\mathcal{N}_{s}} E^s_i + \sum_{i}^{\mathcal{N}_{b}} E^b_i,
\end{equation}
where $\mathcal{N}{s}$ and $\mathcal{N}_{b}$ are the numbers of stretching and bending elements, respectively.
The internal elastic force can be derived as:
\begin{equation}
\mathbf{F}^{\text{ela}} = - \frac {\partial E^{\text{ela}}} {\partial \mathbf{q} },
\end{equation}
which will be used later to construct the equation of motions of the simulated robot.

\subsection{Frictional contact force}
In this section, we discuss the computation of frictional contact forces between a soft robot and a rigid substrate.
When a node $\mathbf{x}_{i}$ makes contact with the ground, frictional contact force responses are generated between this node and the substrate. The contact distance $d_i$ and the contact normal force $(\mathbf{F}^{\mathrm{con}}_{n})_{i}$ must satisfy Signorini's condition: $\lVert (\mathbf{F}^{\mathrm{con}}_{n})_{i} \rVert \geq 0 \perp 0\leq d_i$. To incorporate this constraint into the numerical framework, a smooth log-barrier function is constructed:
\begin{equation}
(\mathbf{F}^{\mathrm{con}}_{n})_{i} = K_{c} \left[ - 2 (d_{i} - \tilde{d}) \log( \frac {d_{i}} {\tilde{d}}) - \frac {(d_{i} - \tilde{d})^2} {d_{i}} \right] \mathbf{n}_{i},
\label{eq:normalContactF}
\end{equation}
where $d_{i} = \mathbf{x}_{i} \cdot \mathbf{n}_{i}$ is the distance to the ground, $\tilde{d}$ is the barrier parameter, and $K_{c}$ is the contact stiffness. The contact force vanishes when $d_i$ is larger than $\tilde{d}$ and it increases as $d_{i}$ approaches zero. 

For the friction, we utilize Coulomb's friction law. When the velocity of a contact node along the tangential direction is non-zero, defined as:
\begin{equation}
\mathbf{v}_{i}^{t} =\mathbf{v}_{i} - \mathbf{v}_{i} \cdot \mathbf{n}_{i} \neq \mathbf{0},
\end{equation}
a smooth frictional force is applied based on the maximum dissipative principle. For small relative velocities, $0 \le ||\mathbf{v}_{i}^{t}|| < \epsilon_{v}$, the tangential frictional force is
\begin{equation}
(\mathbf{F}^{\mathrm{con}}_{t})_{i} =
- \mu || (\mathbf{F}^{\mathrm{con}}_{n})_{i} || \frac {\mathbf{v}_{i}^{t}} {||\mathbf{v}_{i}^{t}|} \left(- \frac {||\mathbf{v}_{i}^{t}||^2} {\epsilon_{v}^2} +  \frac {||\mathbf{v}_{i}^{t}||} {\epsilon_{v}} \right),
\label{eq:frictionF1}
\end{equation}
where $\mu$ is the friction coefficient and $\epsilon_{v}$ is the normalization parameter. For larger velocities $||\mathbf{v}_{i}^{t}|| \geq \epsilon_{v}$, the tangential frictional force is
\begin{equation}
(\mathbf{F}^{\mathrm{con}}_{t})_{i} =
- \mu || (\mathbf{F}^{\mathrm{con}}_{n})_{i} || \frac {\mathbf{v}_{i}^{t}}  {|| \mathbf{v}_{i}^{t} ||}.
\label{eq:frictionF2}
\end{equation}
Introducing $\epsilon_{v}$ smooths the traditional Heaviside frictional force, making the transition from static to sliding friction differentiable and solvable through gradient-based methods. With Eqs.~(\ref{eq:normalContactF}),~(\ref {eq:frictionF1}) and~(\ref{eq:frictionF2}), the frictional contact responses can be fully expressed in terms of the DOF vector $\mathbf q$, which can subsequently be used to construct the equation of motions for the simulated robot.

\subsection{Time marching scheme}
The DOF vector $\mathbf{q}$ is updated from the current time step ($t_{k}$) to the next time step ($t_{k+1}$) using the classical implicit Euler method for its simplicity and robustness. At the $k$-th time step, $t_{k}$, with known DOF vector $\mathbf{q}(t_{k})$ and velocity $\dot{\mathbf{q}}(t_{k})$, the equations of motion are updated as follows
\begin{equation}
\begin{split}
{\mathbb{M}}  { \ddot{\mathbf{q}}(t_{k+1})} &= \mathbf{F}^{\text{ela}}(t_{k+1}) + \mathbf{F}^{\text{gra}}(t_{k+1}) + \mathbf{F}^{\text{con}}(t_{k+1}), \\
\mathbf{q}(t_{k+1}) &=  \mathbf{q}(t_{k}) + \Delta t \dot{\mathbf{q}}(t_{k+1}), \\
\dot{\mathbf{q}}(t_{k+1}) &=  \dot{\mathbf{q}}(t_{k}) + \Delta t \ddot{\mathbf{q}}(t_{k+1}),
\end{split}
\label{eq:implicitEuler}
\end{equation}
where $\mathbb M$ is the lumped mass matrix of the system, $\Delta t$ is the time step size, and $\mathbf{F}^{\text{gra}}$ is the gravitational force. The iterative Newton-Raphson method is used to solve the nonlinear equations of motion and update $\mathbf q$ via time marching.

\section{Dynamic Analysis of Jump}
\label{sec:analysis}
In this section, we utilize our developed simulation tool to examine the jump process of the snap-actuated robot. The jumping process includes two stages: i) the initial actuation of the bi-stable beam, and ii) the subsequent rigid contact interactions between the beam and rigid substrate. We characterize the jump dynamics and provide a benchmark for designing the snap-actuated robot.

\subsection{Snap-through of elastic beam}
The pre-compressed beam undergoes snap-through buckling by rotating the boundary, as illustrated in Fig.~\ref{fig:mechanicsPlot}(b1), which generate the inversion and can be adopted as the actuator of the robot.
When both ends of the beam rotate simultaneously at the same rate, the beam (initially in the upward position, shown in black) quickly transitions to the opposite stable state (shown in blue). During this snap-through process, the beam passes through a critical state (shown in yellow). This transition releases stored elastic energy, which actuates the robot's jump.

To thoroughly understand this jumping mechanism, we first use our numerical framework to perform a comprehensive static analysis of the beam's buckling behavior. 
In the simulation, we set the number of nodes $\mathcal{N}=100$ and time step size $\Delta t=1$ ms to ensure numerical convergence and accuracy.
For the generality of the analysis, we convert all quantities to be dimensionless, denoting them with $\bar{(\cdot)}$ hereafter.
We choose to monitor the normalized midpoint height $\bar{y}_{{1}/{2}} = y_{{1}/{2}} / L$ and the normalized elastic energy $\bar{E} = E^{\mathrm{ela}} L / EI$ to track the state of the beam as the rotation angle changes. 
A sudden jump in these values occurs when the rotational angle $\alpha$ achieves a certain threshold, which implies that snap-through buckling happens.
Note that, the normalized released energy is controlled by the pre-compression ratio $\varepsilon = \Delta L / L$ of the beam, and such relation is linear, referring to Fig.~\ref{fig:mechanicsPlot}(b3).

Next, we examine how the beam's geometry transitions from one stable configuration to the other, which also determines the robot's jump.
We perform a dynamic analysis to understand this transition.
A fact is observed: if the rotation angles on the two ends are not identical, the transition shifts from symmetric to asymmetric.
We introduce a new quantity $\delta \alpha$ to quantify the difference between the rotation angles at the two ends.
In Figs.~\ref{fig:mechanicsPlot}(c1) and (c2), the dynamic transition of the configuration during the snap-through process is illustrated for two cases: asymmetric clamped boundaries ($\delta \alpha = 0.1)$ and symmetric clamped boundaries ($\delta \alpha = 0.0)$, respectively. The maximum midpoint angle $\theta_\textrm{max}$ during the snap-through, seeing Figs.~\ref{fig:mechanicsPlot}(c1), is used to evaluate the symmetry of the configuration, i.e., the transition is symmetrical when $\theta_\textrm{max} = 0$ and asymmetrical when $\theta_\textrm{max} \neq 0$  \cite{wang2024transient}.
For those two cases, the pre-compression ratio is fixed as $\varepsilon = 0.1$, and the loading rate is $\dot{\alpha} = 20 \; \si{rad/s}$. 
By controlling the mismatch angle $\delta \alpha$, we can manage the dynamic transitions of the beam, which could potentially offer a wider athletic ability for the robot's jump.
By monitoring the maximum midpoint angle $\theta_\textrm{max}$ as a function of the mismatch angle $\delta \alpha$, we can observe a linear increase for smaller $\delta \alpha$ values, but beyond a certain threshold, the mismatch angle has minimal influence on the geometry transitions during the buckling process.

\begin{figure*}[t]
\centering
\includegraphics[width=1.0\textwidth]{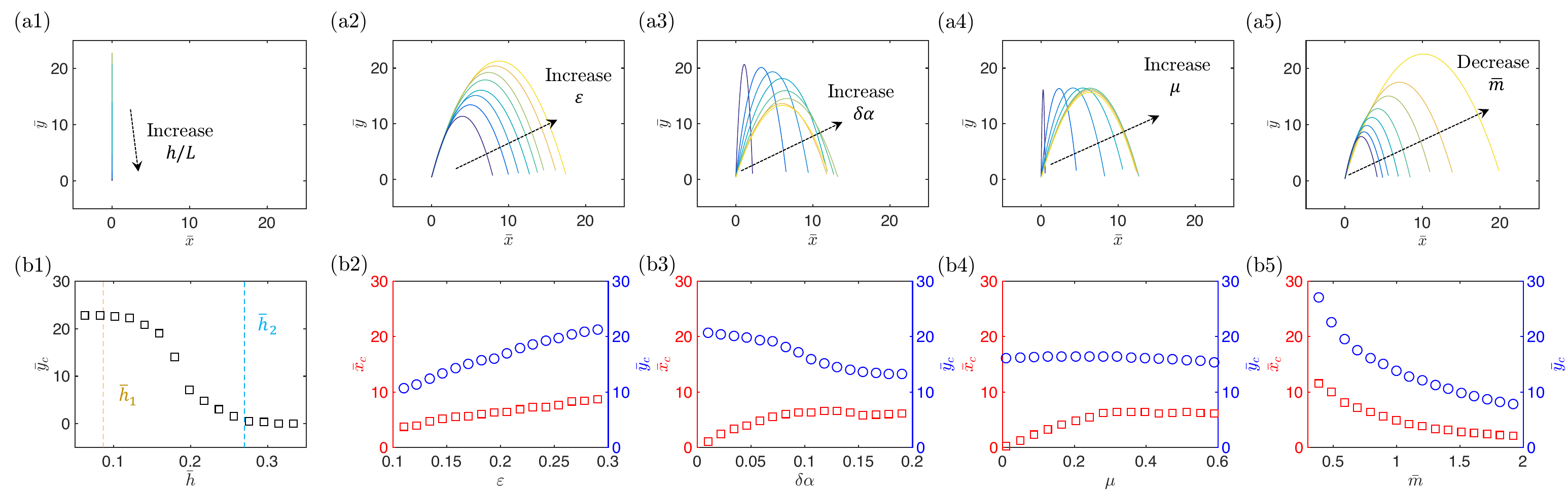}
\caption{(a) Trajectories for jump robot with different design parameters. (b) Maximum jump height and the associated distance, normalized by beam length, as a function of different parameters: (1) beam height, $h$, (2) pre-compression ratio, $\varepsilon$, (3) angle mismatch $\delta \alpha$, (4) frictional coefficient, $\mu$, and (5) normalized mass, $\bar{m}$.} 
\label{fig:jumpPlot}
\end{figure*}

\subsection{Contact-induced jump of robot}
Moving forward, we analyze the jump dynamics of the snap-actuated robot, driven by the interplay between nonlinear elasticity and frictional contact.
Here, we use $\mathcal{N}=190$, $\Delta t=0.01 \si{ms}$, $K_{c} = 10^4 \si{N/m}$, $\tilde{d} = 0.5\si{mm}$, $\epsilon_{v} = 10^{-4} \si{m/s}$ for the numerical investigation through a convergence study.
The geometric and physical parameters of the robot, as detailed in Table~\ref{tab:para}, are consistent with those in the prior work by Wang et al.~\cite{wang2023insect}.
Depending on whether the dynamic transition of the beam's buckling is symmetric or asymmetric, the robot can achieve two distinct types of locomotion: a vertical jump (perpendicular to the rigid substrate) or an inclined jump (oblique to the rigid substrate). To ensure the rigidity of the robotic frame, its stiffness is set to be 1,000 times greater than that of the beam in the numerical simulation.

\begin{table}[ht]
\centering
\caption{Physical and geometric parameters }
\label{tab:para}
\begin{tabular}{cccc}
\toprule
Description & Symbol & Value & Unit \\
\midrule
Beam length & $L$ & $20.0$ & \si{mm} \\ 
Beam width & $w$ & $10.0$ & \si{mm} \\ 
Beam thickness & $b$ & $0.05$ & \si{mm} \\ 
Young's modulus & $E$ & $200.0$ & GPa \\
Material density & $\rho$ & $5000.0$ & \si{kg/m^3} \\
Robot height & $l_{2}$ & $10.0$ & \si{mm} \\
Gravity & $g$ & $-10.0$ & \si{m/s^2} \\
\bottomrule
\end{tabular}
\end{table}

\subsubsection{Optimal installation position of the actuator}
The snap-through dynamic transition of the beam can strike the ground and actuate the robot's jump.
Therefore, in addition to the snap-through buckling of the beam, the mounting position of the actuator — referred to as the beam height $h$ (the distance between the beam's mounting position and the rigid substrate) — is also significant.
In this section, we focus on how the beam height $h$ influences the robot's jump.

We examine a simple vertical jump enabled by a symmetric snap-through, characterized by zero angle mismatch, i.e., $\delta \alpha = 0$. 
In this analysis, $h$ is set as the primary variable while all other parameters remain constant: pre-compression ratio $\varepsilon = 0.2$, total mass $m = 4.0 $ g (normalized mass $\bar{m} = mgL^2/ EI = 0.768$), and frictional coefficient $\mu = 0.0$.
When invoking the snap-through buckling of the beam, the robot will jump vertically due to the strike between the beam and the ground, as shown in Fig.~\ref{fig:jumpPlot}(a1). 
It is noted that the beam height $h$ has a significant influence on the normalized maximum jump height, $y_c/L$. 
In Fig.~\ref{fig:jumpPlot}(b1), it can be observed that the normalized maximum jump height remains relatively constant for small beam height $h$, e.g., $h \le h_{1}$, where $h_{1}$ corresponds to the height at the critical configuration before the snap, as colored yellow in Fig.~\ref{fig:mechanicsPlot}(b1).
As $h/L$ increases, the normalized maximal jump height starts to decrease, which implies the strike between the beam and the ground becomes smaller. This observation aligns with the findings from our previous study \cite{wang2023insect}. When $h \ge h_{2}$, the robot fails to jump because the beam can not make contact with the ground, where $h_{2}$ corresponds to the height at the critical configuration after the snap, as colored blue in Fig.~\ref{fig:mechanicsPlot}(b1).
We use static analysis to derive the relationship between the pre-compression $\varepsilon$ and the best height, $h_{1} = f(\varepsilon)$, i.e., with a prescribed compression, we can derive an optimal height design, $h = h_{1}$.
We chose the optimal beam height since the robot can use the released elastic energies during the buckling process more efficiently.


\subsubsection{Design parameters of the jump}
\label{sec:design_param}
We next investigate the design parameters of the robot's jump.
Here, the normalized beam height is fixed as $\bar h = h_1/L$.
According to the mechanical analysis, four parameters primarily control the robot's jump: pre-compression ratio $\varepsilon$, angle mismatch $\delta \alpha$, frictional coefficient $\mu$, and normalized total mass $\bar m =  mgL^2/EI$.
Therein, a forward model can be constructed to describe the relationship between the four parameters and the robot's jump:
\begin{equation}
(\bar x_c, \bar y_c)  = \mathcal{F}(\delta\alpha, \varepsilon, \bar m, \mu),
\label{eq:forward_model}
\end{equation}
where ($\bar x_c = x_c/L$, $\bar y_c = y_c/L$) is the normalized highest jumping point.
Among these four parameters, the normalized net mass $\bar{m}$ and the friction coefficient $\mu$ are fixed once the robot is fabricated and the substrate is determined; therefore, they are referred to as "environmental parameters." In contrast, the rotation angle mismatch $\delta \alpha$ and the pre-compression ratio $\varepsilon$ can be varied during actuation and are thus referred to as "controllable parameters."
The forward model can be solved using our proposed numerical framework. To examine the influence of each design parameter, we establish baseline parameters as follows: $\varepsilon = 0.2$, $\delta \alpha = 0.1$, $\mu = 0.3$, and $\bar{m} = 0.768$. We then conduct controlled numerical experiments to explore the effects of these parameters.

Based on the experimental setup described in \cite{wang2023insect}, we adjust the frame's mass to fine-tune the total mass of the jumping robot.
The contributions of various design parameters are summarized in Fig.~\ref{fig:jumpPlot}.
First, the pre-compression ratio $\varepsilon$ increases both $\bar{x}_c$ and $\bar{y}_c$ monotonically, as greater energy is released during the snap-through process.
Second, as the angle mismatch $\delta \alpha$ increases, $\bar{x}_c$ increases while $\bar{y}_c$ decreases until they plateau, indicating an optimal range of $\delta \alpha$ for effectively tuning the robot's jumping locomotion.
Next, the friction coefficient $\mu$ does not affect $\bar{y}_c$, since the vertical reaction forces are perpendicular to friction and thus linearly independent. However, $\bar{x}_c$ increases with increasing friction coefficient. There is a threshold for the friction coefficient beyond which it no longer affects the robot's transverse jumping abilities.
Lastly, the normalized mass $\bar{m}$ decreases both $\bar{x}_c$ and $\bar{y}_c$ as the system's inertia increases.
Considering the influence of each design parameter and the fabrication constraints from \cite{wang2023insect}, we selected the following design parameter ranges for the proposed jumping robot: $\varepsilon \in [0.1, 0.3]$, $\delta \alpha \in [0.01, 0.19]$, $\mu \in [0.1, 0.6]$, and $\bar{m} \in [0.3, 2.0]$.

\section{Inverse design}
\label{sec:inverseDesign}

\begin{figure}[ht]
\centering
\includegraphics[width=0.47\textwidth]{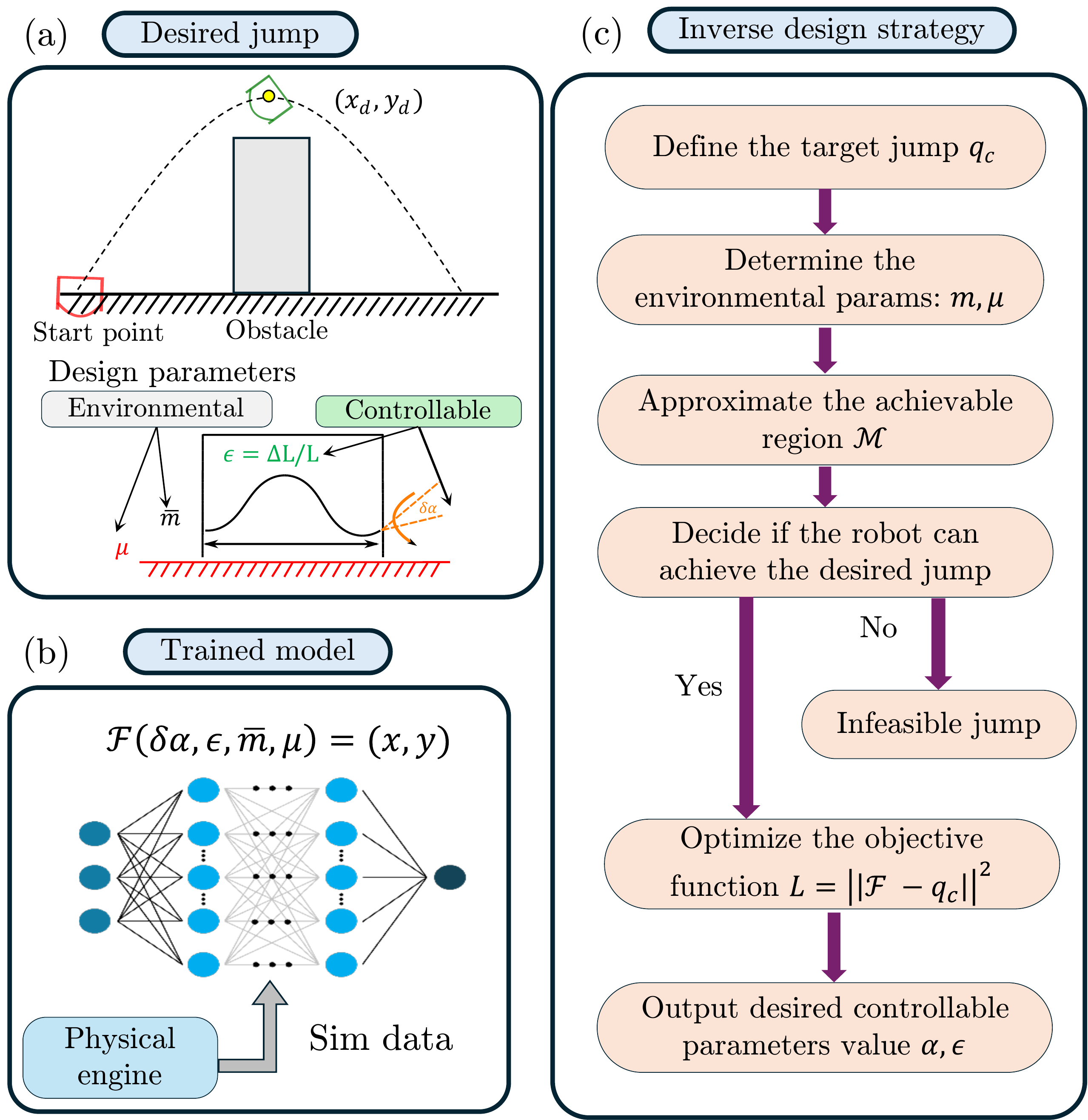}
\caption{Inverse design of the jumping robot. (a) Illustrations of all design parameters, including the desired jump ($x_c^d, y_c^d$), controllable parameters (angle mismatch $\delta\alpha$ and pre-compression ratio $\varepsilon$), and environmental parameters (friction coefficient $\mu$ and normalized mass $\bar m$). (b) A lightweight forward model is trained using simulation data to predict the robot's jump based on the given design parameters. (c) The overall inverse design strategy.}
\label{fig:schematic}
\end{figure}

In this section, we explore how to design the robot's parameters to achieve a desired jump. Although our proposed numerical tool can perform real-time simulations, it remains infeasible to explore all possible design parameter combinations due to the curse of dimensionality. To address this challenge, we learn the forward model in Eq.(\ref{eq:forward_model}) using a data-driven approach. By leveraging efficient parallel computing and the differentiability of the trained forward model, we propose an efficient and accurate inverse design scheme to guide the manufacturing and control of jumping robots for various task requirements. The overall structure of our inverse design process is illustrated in Fig.\ref{fig:schematic}.

\subsection{Data Generation}
To learn the forward model, as illustrated in Fig.\ref{fig:schematic}(b), we first utilize our developed simulation tools to generate numerous sampled parameter combinations $(\delta \alpha, \varepsilon, \bar m, \mu)$ and their corresponding highest points $(\bar x_c, \bar y_c)$. This process resulted in a training dataset $\mathcal{D}$ consisting of six-element tuples $(\delta\alpha, \varepsilon, \bar m, \mu, \bar x_c, \bar y_c)$.
To streamline the generation of training data, we sample the dataset within the design parameter space defined in Sec.\ref{sec:design_param}.
Using our simulation framework, we generate a dataset $\mathcal{D}$ containing a total of 50,625 training samples, which required approximately 4 hours of computing time on an Intel i9-12900H processor.

\subsection{Learning a forward model}
To learn the forward model $\mathcal{F}$, we use a simple fully-connected feed-forward non-linear regression network with three hidden layers, each with 372 units, to represent the achievable region $\mathcal{M}$. Each layer, except the final output layer, is followed by a rectified linear unit (ReLU) activation. Additionally, we standardize all inputs during pre-processing as
\begin{equation}
\mathbf i' = \frac{\mathbf i - \bar{\mathbf i}_D}{\boldsymbol{\sigma}_D},
\end{equation}
where $\mathbf i$ is the original input, $\bar{\mathbf i}_D$ the mean of the dataset $\mathcal{D}$, and $\boldsymbol{\sigma}_D$ the standard deviation of $\mathcal{D}$.

We perform an initial $80-20$ train-validation split on the dataset $\mathcal{D}$ with a batch size of 64. Mean absolute error (MAE) is used as the training error metric. We train the model using the Adam optimizer and gradually increased the batch size up to 1024. This approach enable us to achieve an MAE of less than $7\times10^{-2}$.

\subsection{Methodology for the inverse design}
\begin{figure}[ht]
\centering
\includegraphics[width=0.75\columnwidth]{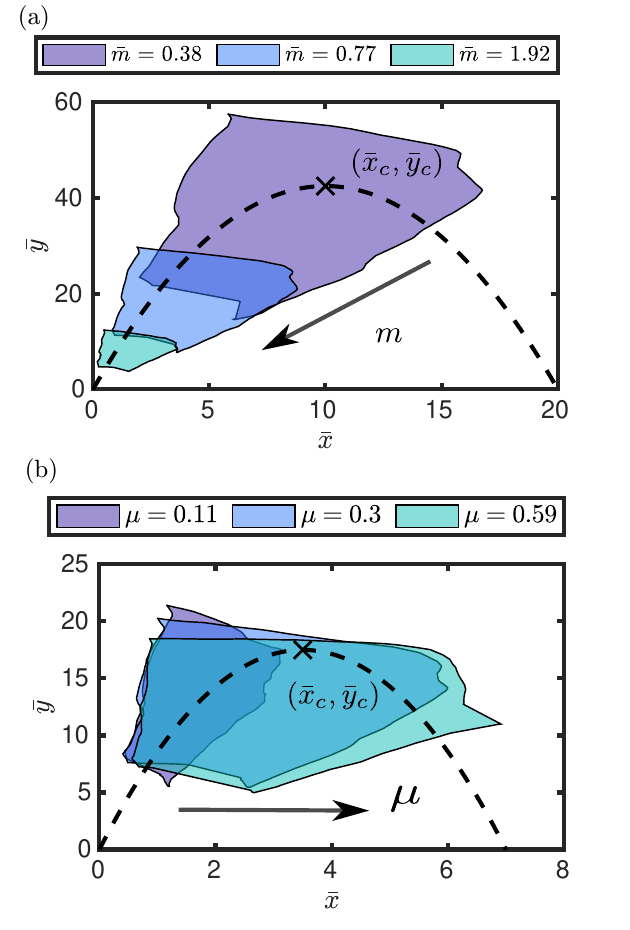}
\caption{Influence of environmental parameters on the available region $\mathcal{M}$. (a) Influence of the normalized robot mass $\bar m$ with a constant friction coefficient $\mu = 0.3$. (b) Influence of the friction coefficient with a constant robot mass $\bar m  = 0.768$.}
\label{fig:planning_region}
\end{figure}

The inverse design of the jumping robot is to explore the combination of the controllable parameters: the pre-compression ratio $\varepsilon$, angle mismatch $\delta\alpha$, along with the environmental parameters including the normalized mass $\bar m$ and the surface friction coefficient $\mu$, to achieve a target jump. It is worth noting that the achievable region $\mathcal{M}$ is finite, as shown in Fig.~\ref{fig:planning_region}, given the manufacturing limitation and actuation mechanism of the robots. 

Therein, the inverse design of the robot can be divided into two steps. First, we determine whether the robot can reach the target point. Second, we explore the values of the controllable parameters based on the specific environmental factors.

With the efficient parallel computing capabilities of the trained model, we can estimate the achievable region of the jumping robot with the given combination of environmental parameters. 
We discretize the controllable parameter space into rectangular grids consisting of $\delta_\alpha \times \delta_\varepsilon$ blocks, where $\delta_\alpha$ and $\delta_\varepsilon$ are the sampling distance of the controllable parameters.
For the midpoint of each block, we compute the corresponding $(\bar x_c, \bar y_c)$, resulting in a $2 \times m \times n$ matrix that approximates the achievable region $\mathcal{M}$, as shown in Fig.~\ref{fig:planning_region}.
Once $\mathcal{M}$ is determined, a ray-casting algorithm is used to decide if the target point is reachable. 
If achievable, an objective cost function is constructed to evaluate the design parameters using the trained model
\begin{equation}
J(\delta\alpha, \varepsilon) = \lVert \mathcal{F}(\delta\alpha, \varepsilon, \bar m, \mu) - (x_c^d, y_c^d) \rVert^2.
\label{eq::loss}
\end{equation}

The gradient of the cost function can be obtained using the chain rule
\begin{equation}
\begin{aligned}
\nabla J &= [\frac{\partial J}{\partial (\delta \alpha)}, \frac{\partial J}{\partial \varepsilon}],\\
\frac{\partial J}{\partial (\delta\alpha)} &= \frac{\partial \mathcal{F}}{\partial (\delta\alpha)} \left( \mathcal{F}(\delta\alpha, \varepsilon, \bar m, \mu) - (\bar x_c^d, \bar y_c^d) \right),  \\
\frac{\partial J}{\partial \varepsilon} &=  \frac{\partial \mathcal{F}}{\partial \varepsilon} \left(\mathcal{F}(\delta\alpha, \varepsilon, \bar m, \mu) - (\bar x_c^d, \bar y_c^d) \right). \\
\end{aligned}
\label{eq::gradloss}
\end{equation}

\begin{figure}[ht]
\centering
\includegraphics[width=0.75\columnwidth]{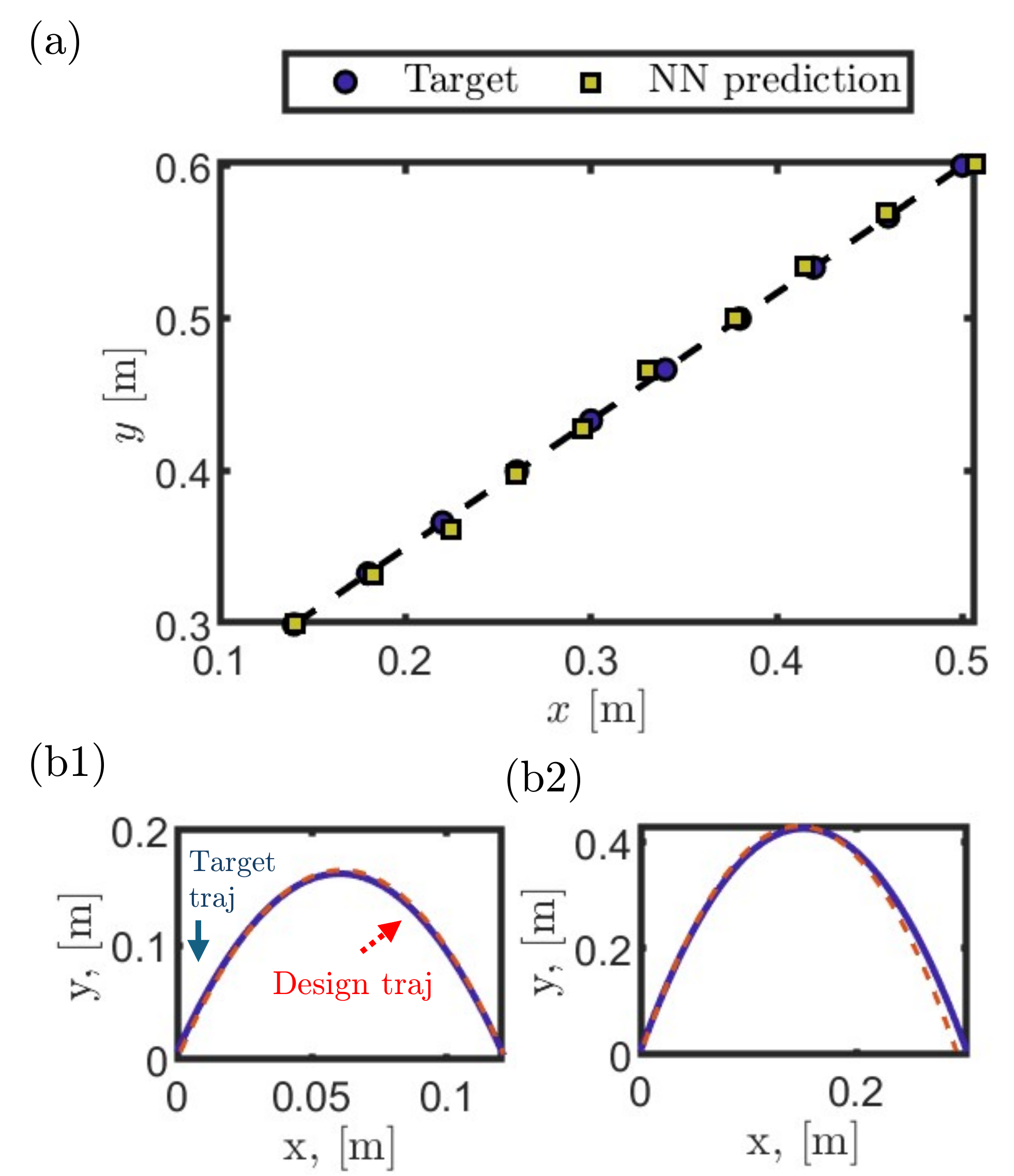}
\caption{(a) The inverse design of a specific robot to achieve multiple desired jumping points. (b) The comparison between the designed robot trajectory and the target trajectory for jumping robots under different environmental parameters.}
\label{fig:results}
\end{figure}

With Eq. (\ref{eq::gradloss}), we employ Adam optimizer to find the optimal controllable parameters minimizing the cost function defined in Eq. (\ref{eq::loss}). 
This inverse design approach enables us to determine the design parameters required to achieve the desired jump.
These parameters are subsequently used in numerical simulations to validate the effectiveness of our proposed method.
Fig.~\ref{fig:results} presents parts of the validation results for the proposed inverse design strategy.
The predicted design parameters successfully control the robot to follow the desired jump trajectory with excellent performance. 

We further validate our inverse design strategies using 100 randomly generated combinations of environmental parameters, representing a variety of robots and scenarios.
Comparing the jumps actuated by the solved inverse design parameters with the target jumps, we observe an average error of $0.0029 $ m and a standard deviation of $0.0025 $ m. Additionally, the average computation time for our proposed strategy is $0.33 $ s.
Based on these evaluation metrics, we conclude that the proposed inverse design scheme is efficient and accurate. Due to its lightweight nature and high computational speed, it can be implemented onboard a jumping robot for real-time control.

\section{Conclusion}
\label{sec:conclusion}

In this study, we introduced a novel design for a jumping robot that leverages a tunable snap-through buckling mechanism to achieve a controllable jump. We thoroughly analyzed the snap-buckling actuation mechanism and its relationship with the design parameters using a DDG-based reduced-order numerical framework. A physical engine built upon this framework vividly simulates the jumping robot's performance under different scenarios. Using data from these simulations, we trained a light-weight data-driven model that accurately infers the optimal design parameters for achieving a desired jump. Our inverse design scheme was validated against the physical engine, demonstrating the ability to plan jumps with millimeter-scale error in less than $0.5 \si{s}$. 

This work sheds light on the design of soft jumping robots by leveraging the structural buckling power. The simplicity, accuracy, and high computational efficiency of our inverse design scheme make it well-suited for installation on untethered jumping robots for online control. Future research will focus on manufacturing such an untethered jumping robot with a controller based on the inverse design strategy. By integrating the efficient, accurate data-driven model with onboard power and micro-controllers, we aim to fabricate a jumping robot capable of navigating and adapting to complex environments without external intervention.

\section*{Acknowledgment}

M.L. acknowledges support via start-up funding from the University of Birmingham, UK. W.H. acknowledges the start-up funding from Newcastle University, UK.

\appendices

\bibliographystyle{IEEEbib}
\bibliography{robot}

\end{document}